\documentclass[11pt,a4paper]{article}
\usepackage[T1]{fontenc}
\usepackage[utf8]{inputenc}
\usepackage{lmodern}
\usepackage[margin=24mm]{geometry}
\usepackage{microtype}
\usepackage{graphicx}
\usepackage{amsmath,amssymb}
\usepackage{booktabs,longtable,array}
\usepackage[numbers,sort&compress]{natbib}
\usepackage{xurl}
\usepackage[hidelinks]{hyperref}
\usepackage{bookmark}
\usepackage{needspace}
\newlength{\atlwidth}
\newcommand{\doi}[1]{\href{https://doi.org/#1}{\nolinkurl{doi:#1}}}
\hypersetup{pdftitle={AI to Learn 2.0: From AI-Assisted Performance to Human Understanding and Durable Knowledge},pdfauthor={Seine A. Shintani},pdfsubject={AI-assisted learning and research; proposed maturity rubric}}
\title{AI to Learn 2.0: From AI-Assisted Performance to Human Understanding and Durable Knowledge}
\author{Seine A. Shintani$^{1,2,3}$\\[5pt]
\small $^1$Department of Biomedical Sciences, College of Life and Health Sciences,\\
\small Chubu University, Kasugai, Aichi 487-8501, Japan\\[3pt]
\small $^2$Center for Mathematical Science and Artificial Intelligence,\\
\small Chubu University, Kasugai, Aichi 487-8501, Japan\\[3pt]
\small $^3$Institute for Advanced Research, Nagoya University,\\
\small Nagoya, Aichi 464-8601, Japan\\[4pt]
\small Correspondence: \href{mailto:s-shintani@fsc.chubu.ac.jp}{s-shintani@fsc.chubu.ac.jp}}
\date{}
\begin{document}
\maketitle
\begin{abstract}
How can working with AI help a person understand a problem well enough to tackle a new one? I propose AI to Learn 2.0 as a way to pursue that goal in research and education. Alongside performance on the assisted task, I consider what the person can later explain or do under stated conditions (the capability residual) and what resources they can keep, examine and revise (the artifact residual). A binding model illustrates the distinction. Two sets of rates predict the same amount bound at equilibrium. Understanding why they agree suggests when to measure a difference. In my research on heated heart muscle cells, equations suggested by symbolic regression prompted me to examine neighbouring contractile units. Measurements and nonlinear analyses showed how a nearly constant period can coexist with chaotic local dynamics. I also used AI to develop a study system covering 21 rounds of Japan's national examination for clinical laboratory technologists. Students can revisit missed questions, and instructors can revise explanations in light of common errors. Routine practice uses stored materials without calling generative AI. I describe a five-part package for retaining a result and its supporting evidence, seven dimensions for reviewing it and a separate scale for evidence of human capability. The supplied code and offline review sheets agree on tests of the decision rules. Whether this approach improves learning or helps groups sustain expertise remains an empirical question.
\end{abstract}
\Needspace{6\baselineskip}\section{Introduction}

When I use artificial intelligence (AI) in research or teaching, I want to come away able to explain something I could not explain before. AI might suggest two models that fit the observations equally well. To choose my next measurement, I need to work out where their predictions diverge. A student who solves a problem with an AI-generated hint faces a similar challenge when one condition in the problem changes. What can the student now work out?

In application-oriented modeling, a researcher may ask how accurately a model predicts an outcome over the range of intended use. In understanding-oriented modeling, the question may be which relations give rise to the outcome and what would happen if a condition changed. A single mathematical model can serve both aims \citep{Breiman2001,Shmueli2010}. A model with little explanatory detail can be valuable for prediction. An explicit relation can suggest a new experiment whose outcome could support or challenge the proposed explanation \citep{Krenn2022,Messeri2024}. In education, too, the evidence depends on the question. A correct answer during practice measures performance on that occasion. Solving an unfamiliar problem later can reveal what the student learned \citep{SoderstromBjork2015}.

AI systems can propose explanations, write programs and call tools to carry out parts of a research task \citep{Yao2023,Gottweis2026}. A fluent explanation may contain a claim with no supporting evidence \citep{Ji2023}. Even a correct calculation can answer a different question from the one the researcher intended. As the range of tasks expands, I find it useful to decide which steps to delegate and which judgments I want to learn to make myself.

\begin{figure}[htbp]
\centering
\includegraphics[width=\linewidth]{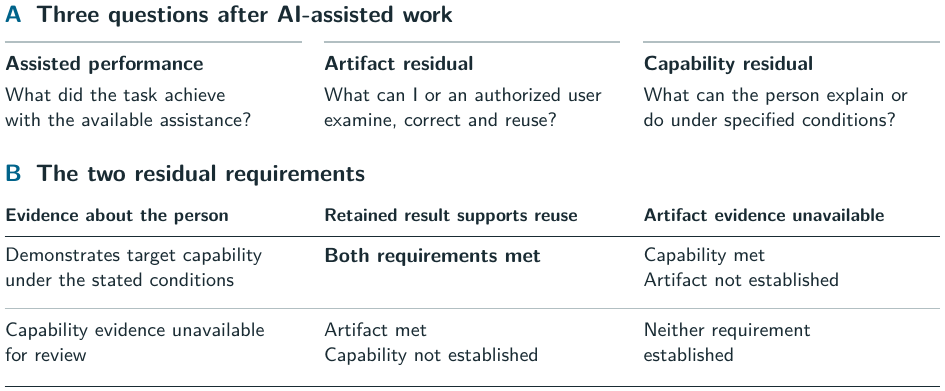}
\caption{Three outcomes of AI-assisted work. (A) Immediate performance, resources retained for later work, and evidence of a person's capability. (B) Four illustrative combinations of the two residual requirements. Evidence supporting one requirement does not establish the other. Section~\ref{sec:rubric} distinguishes a known deficiency from missing evidence.}
\label{fig:outcomes}
\end{figure}

I originally proposed AI to Learn as a way to use AI whose workings are difficult to inspect and still obtain knowledge and tools that remain useful without returning to it \citep{ShintaniAItoLearn2023,Shintani2025}. In AI to Learn 2.0, I ask two questions about what remains after a task \citep{Shintani2026}. What can the person explain or do when asked to demonstrate the relevant skill? This is the \emph{capability residual}. What can the person keep using, check against the evidence, or revise for a new purpose? This is the \emph{artifact residual}. The same piece of work can support either or both (Figure~\ref{fig:outcomes}). Running a program gives a result; choosing how to change it for a new experiment provides a separate opportunity to demonstrate understanding.

A researcher who keeps a model, the measurements and the reasons for choosing them can return to a question months later. In a laboratory the same records can help a colleague learn the method and carry it forward when the original researcher leaves. Unpublished observations, for example, may be available only to authorized reviewers. What the group preserves and who can examine it depend on the aim of the work and the nature of the information. I develop these ideas below through examples from my research and teaching.

\Needspace{6\baselineskip}\section{Why learning through AI matters}

\subsection{Prediction, explanation and scientific understanding}

The purpose of a model affects how to build and evaluate it. Breiman compared modeling the process that generates data with building an algorithm for prediction. Shmueli examined how explanatory and predictive aims shape choices about variables, fitting and evaluation \citep{Breiman2001,Shmueli2010}. Computational systems can help researchers find relations that make a phenomenon easier to understand. They can also produce more findings than researchers can critically examine \citep{Krenn2022,Messeri2024}. Evaluating a model's output and asking what its user understands therefore call for different evidence.

Protein structure prediction illustrates the scientific value of an accurate prediction \citep{Jumper2021}. Computational searches can also yield explicit constructions or candidate equations \citep{Romera2024,Udrescu2020,Udrescu2020b,Cranmer2020}. A proposed equation invites two further questions. Under what assumptions does it hold, and what observation could distinguish it from another equation that fits the same data? Verification connects a result to the claim it can support \citep{Cornelio2026}. The binding model in Section~\ref{sec:binding} gives a worked example.

\subsection{Performance, practice and learning}

A student might delegate arithmetic to a tool and concentrate on explaining why an equation applies. Research on the use of external aids, on learning by explaining a solution to oneself, and on the interpretation of assessment results informs this choice \citep{RiskoGilbert2016,Chi1989,Kane2013}. When a tool can solve the entire exercise, a separate task may be needed to find out what the student learned \citep{SoderstromBjork2015}.

In a high-school mathematics study, students using a general-purpose AI interface performed better during practice and worse on a later unaided examination than students who practised without AI \citep{Bastani2025}. Undergraduates studying physics with an AI tutor designed for particular lessons scored higher on the immediate posttest than peers in an active-learning class \citep{Kestin2025}. In a randomized study of programmers learning an unfamiliar library, the AI-assisted group scored lower on a subsequent quiz \citep{ShenTamkin2026}. Beyond the classroom, knowledge workers surveyed about AI use described directing effort toward checking and integrating its responses \citep{Lee2025}. Each study concerns particular tasks and conditions. To assess what a learner takes from an activity, it is useful to specify what they should be able to do later and to measure it over time. A recent research announcement outlines one such measurement effort \citep{OpenAILearning2026}.

\subsection{Resources for the next inquiry}

An explicit equation or an editable program gives its user something to return to. The next time an assumption changes, the user can revise a step and check the consequences. Work on reusable research software and on the distinction between using AI to develop an analysis and using it to run that analysis informs this practice \citep{Barker2022,Dunn2026}. A review of reliance on AI and a report based on interviews with firms also raise the question of how groups sustain the expertise needed to judge results \citep{DuYuan2026,Wilmers2026}. A report from a research pilot documented scientists using AI to develop research ideas \citep{Liao2026}.

For a particular task, I start by identifying a judgment or skill worth developing and a resource worth keeping. I can then choose a way to check each. Methods for interpreting models, documenting models and data, and keeping people responsible for decisions complement this approach \citep{Rudin2019,Mitchell2019,Gebru2021,Perkins2024,Santoni2018}. The next sections set out what evidence to retain and how to use it in a review.

\Needspace{6\baselineskip}\section{The review object and the two residual requirements}

\Needspace{4\baselineskip}\subsection{Specify the work before scoring it}

Begin with the task and the knowledge or capability worth developing. An annotated derivation, for example, can be examined for the correctness of the mathematics. If a learner submits it as evidence of understanding, the reviewer must also consider what that learner can explain. Specify the purpose, the claims that matter for that purpose, the intended users (who may include the creator), the point of review, the tools or resources the work depends on, and any capability being assessed. I denote this context by $\Gamma$.

Record immediate performance separately using task-appropriate outcomes and the actual assistance conditions, when relevant. Retain the adopted question or procedure and the evidence used to evaluate it, including tool contributions material to the claim. The record can be private where appropriate. For example, a binding curve may predict the equilibrium occupancy even though the speed of binding remains unknown. The equation and its derivation can support that limited use. Establishing that a learner can choose the next measurement requires an observable decision by the learner under stated conditions.

\Needspace{4\baselineskip}\subsection{Artifact residual}\label{sec:artifact}

\textbf{Definition 1 (Artifact residual).} For the declared routine use, the intended user or reviewer must be able to use, examine and revise the retained artifact and its accompanying evidence without invoking the opaque AI used to develop it. The grounds for each important claim must be checkable independently of that AI's assurance. This independence concerns the originating AI and the declared use. The user may rely on ordinary software and relevant expertise.

AI can stimulate an investigation whose final product is an independently developed analysis, explanation or procedure. Review follows the evidence supporting that product, including the role of the initial proposal where it matters.

Further AI assistance can be valuable throughout reuse and revision. Two forms of dependence deserve separate attention: operational dependence (an AI component is needed to execute the work) and evidential dependence (the claim rests on the originating AI's assurance). A conventional program written with AI requires evidence of correctness even when it runs independently. A prediction system that uses AI at runtime can support a bounded claim through external testing. Report both forms of dependence.

Before judging independence, identify what is to be used again. The first review dimension, residual opacity ($m_1$ in Section~\ref{sec:rubric}), applies to a retained result. For a live generative service, a reviewer can assess an archived output, a test dataset or a documented performance claim and record which tasks still require the service. Assessing the service itself would require a broader instrument that addresses both continued operation and independent support for its claims.

A model on a local computer can be difficult to interpret, while a script developed with an external service can be inspected and run independently. When a language model consults an external source or uses a tool \citep{Yao2023}, even an accurate calculation may concern the wrong quantity. To check the result later, keep the sources and inputs actually used, the decisive calculations and the decisions that shaped the claim. A short record of these steps can serve this purpose without retaining private deliberations or full conversations.

\Needspace{4\baselineskip}\subsection{Capability residual}

\textbf{Definition 2 (Capability residual).} When a task aims to develop, demonstrate or rely on a human capability, it requires evidence that the designated person or team can give the relevant explanation, make the judgment, adapt the work or recover from a failure. State the task, success criteria, permitted assistance and timing so the observed performance can be attributed to that person or team. Assess the quality of the artifact separately.

For example, asking a script's maintainer to change its units can reveal whether they understand the conversion. Explaining why two equilibrium curves cannot distinguish rate constants tests a different capability from reproducing the curves. Choose aids that leave the reasoning being assessed to the person. Ordinary references, calculators, accessibility aids and collaboration may all be compatible with that aim.

A demonstration establishes what a person can do at the time of assessment. Retention requires a later assessment, growth requires comparison with earlier performance, and a causal claim about AI requires a suitable comparison design. The capability ladder records the strength of the evidence for the stated capability. In self-directed study, it can help learners decide what to practise next.

\Needspace{4\baselineskip}\subsection{The five-part deliverable package}

\begin{equation}\label{eq:package}
D=(A,P,V,F,R).
\end{equation}

Table~\ref{tab:P030} defines five elements that help the person who created a result, or an authorized future user, examine, use and develop it. A short research note or README can contain all five. Choose the level of detail according to the task and the consequences of an error. Capability evidence can be linked to the package and stored separately when, for example, a learner's assessment is private.

\begingroup
\small
\setlength{\tabcolsep}{4pt}
\setlength{\atlwidth}{\dimexpr\textwidth-4\tabcolsep\relax}
\begin{longtable}{@{}>{\raggedright\arraybackslash}p{0.2500\atlwidth}>{\raggedright\arraybackslash}p{0.2600\atlwidth}>{\raggedright\arraybackslash}p{0.4900\atlwidth}@{}}
\caption{The deliverable package and the information needed for review.}\label{tab:P030}\\
\toprule
\textbf{Element} & \textbf{Purpose} & \textbf{What the retained record provides}\\
\midrule
\endfirsthead
\multicolumn{3}{l}{\small Table \thetable\ (continued)}\\
\toprule
\textbf{Element} & \textbf{Purpose} & \textbf{What the retained record provides}\\
\midrule
\endhead
\midrule
\multicolumn{3}{r}{\footnotesize Continued on next page}\\
\endfoot
\bottomrule
\endlastfoot
A: Artifact & Identify the version under review. & Versioned explanation, equation, code, educational resource, or research procedure.\\[3pt]
P: Provenance and verification & Expose the grounds for acceptance. & Sources; operations and decisions material to the claim; checks and revisions supporting accepted claims.\\[3pt]
V: Validity domain & Bound the intended claim and use. & Purpose, assumptions, relevant regime or learners, exclusions, and remaining uncertainties.\\[3pt]
F: Failure and escalation & Make correction possible. & Observable warning signs; when to pause, recheck, correct or withdraw; responsible role.\\[3pt]
R: Operational resources & Make reuse feasible. & Required software, access and costs; retained files; backup, handoff and interruption arrangements.\\[3pt]
\end{longtable}
\endgroup

Choose a check that could reveal a substantive error in the claim. A test generated from a program's mistaken formula can reproduce the same mistake, and several models may cite the same unsupported source. A boundary case checked by hand, a held-out measurement, a separately derived relation or a qualified reviewer's judgment can supply independent grounds. Data leakage and inappropriate evaluation designs are familiar problems in machine-learning-based science \citep{Kapoor2022}. Record what each check supports, any unresolved point, and any limit it places on the claim.

The contents of P depend on the work. For a research analysis, retain the data transformations, model choices, comparisons and validation that support the result. For educational materials, retain the source checks, substantive revisions and decision to release the material. In both cases, a later reviewer needs the grounds for using what was produced.

\Needspace{6\baselineskip}\section{The maturity rubric and its decision rules}\label{sec:rubric}

\Needspace{4\baselineskip}\subsection{Seven indicators of a usable learning or research product}

The rubric rates how ready an artifact is for its declared use. Table~\ref{tab:P036} poses one question for each of seven dimensions and describes what earns a score of 3. Such a description is called a scoring anchor. The technical appendix (Section~\ref{sec:anchors}) gives all 35 anchors for scores from 0 to 4. For each rating, identify the evidence behind it, such as a source, a test result, a versioned record or an observed performance.

\begin{equation}\label{eq:profile}
M=(m_1,\ldots,m_7),\qquad m_i\in\{0,1,2,3,4\}\cup\{U\}.
\end{equation}

\begingroup
\small
\setlength{\tabcolsep}{4pt}
\setlength{\atlwidth}{\dimexpr\textwidth-4\tabcolsep\relax}
\begin{longtable}{@{}>{\raggedright\arraybackslash}p{0.2300\atlwidth}>{\raggedright\arraybackslash}p{0.3600\atlwidth}>{\raggedright\arraybackslash}p{0.4100\atlwidth}@{}}
\caption{The seven review dimensions and their level-3 anchors.}\label{tab:P036}\\
\toprule
\textbf{Dimension} & \textbf{Review question} & \textbf{Level-3 anchor}\\
\midrule
\endfirsthead
\multicolumn{3}{l}{\small Table \thetable\ (continued)}\\
\toprule
\textbf{Dimension} & \textbf{Review question} & \textbf{Level-3 anchor}\\
\midrule
\endhead
\midrule
\multicolumn{3}{r}{\footnotesize Continued on next page}\\
\endfoot
\bottomrule
\endlastfoot
m1: Residual opacity & Can the retained result be used and justified without returning to the originating opaque AI? & Core operation and justification are independent of that AI; noncritical gaps are identified.\\[3pt]
m2: Human authority & Who can accept, correct, restrict or withdraw the result? & A responsible person or team and feasible review, release and correction routes are documented.\\[3pt]
m3: Traceability & Can the operations and evidence material to the accepted claims be reconstructed? & Sources, relevant tool roles, transformations, selection decisions and checks are recorded.\\[3pt]
m4: Validity and failure & Where does the result apply, and what happens when it fails? & Both the scope and observable reconsideration or withdrawal triggers are specified.\\[3pt]
m5: Information protection & Was information access and use appropriate to the task? & Data classes, destinations, minimization, access and relevant retention conditions are documented.\\[3pt]
m6: Resource proportionality & Can the intended work be sustained with the stated resources? & Routine requirements, costs or estimates, and a feasible interruption response are documented.\\[3pt]
m7: Reuse and handoff & Can another appropriate person work from the retained package? & Instructions and files suffice for use and modification without relying on the creator's recollection.\\[3pt]
\end{longtable}
\endgroup

Scores are ordinal. In broad terms, 0 indicates an established absence of the required practice, 1 a weak or merely asserted arrangement, 2 a partial arrangement, 3 a documented arrangement sufficient for the stated use, and 4 an arrangement exercised through an appropriate test or independent use. The dimension-specific wording controls the rating. Unknown evidence is marked $U$, not silently replaced by zero or a presumed pass. An unknown may become a low or high rating when evidence is obtained.

Several dimensions address different aspects of the same task. A script may run independently of its originating model (part of m1), yet its data transformations may be difficult to trace (m3) and another researcher may lack the files needed to reuse it (m7). The authority to correct the script (m2), the evidence for a maintainer's capability, and the resources for continued use (m6) each call for separate records. For $m_4$, rate validity of scope ($m_V$) and failure response ($m_F$) separately. When both are known, $m_4=\min(m_V,m_F)$, so a well-defined scope cannot compensate for an inadequate failure response.

\Needspace{4\baselineskip}\subsection{Critical conditions cannot be averaged away}

Four dimensions must each reach a minimum before the work can be accepted for its proposed use. I retain the four conditions from v1: the retained result can be used and justified independently of the originating opaque AI (m1), a person or team can correct or withdraw it (m2), its scope and the response to failure are documented (m4), and information is protected appropriately (m5). Together these conditions form the critical gate. With all four ratings known, its decision is

\begin{equation}\label{eq:gate}
\operatorname{Gate}(M)=1\quad\Longleftrightarrow\quad\min\{m_1,m_2,m_4,m_5\}\geq 3.
\end{equation}

A known critical score below 3 gives \emph{not met}. For $m_4$, either component below 3 establishes failure even when the other is unknown and $m_4$ itself is $U$. If there is no known failure but a critical rating is unknown, the gate is \emph{not established}. All four critical dimensions must reach 3 for the gate to be \emph{met}. The threshold requires a documented arrangement sufficient for the context. Its empirical calibration remains open.

After checking the gate, review the complete package against its declared purpose. Every material claim needs supporting evidence, and conflicting evidence needs to be resolved. Missing provenance or unusable handoff instructions may prevent acceptance if they are material to the declared use. The m7 rating assesses readiness for handoff; it is particularly relevant to shared or long-lived resources. A private learning resource need not be released to the public. For a fully rated package, a descriptive sum can help compare versions of the same work. The sum has no interpretation as a measure of truth, understanding or safety and carries no certification threshold.

\Needspace{4\baselineskip}\subsection{The separate capability-evidence ladder}

\begingroup
\small
\setlength{\tabcolsep}{4pt}
\setlength{\atlwidth}{\dimexpr\textwidth-4\tabcolsep\relax}
\begin{longtable}{@{}>{\raggedright\arraybackslash}p{0.1400\atlwidth}>{\raggedright\arraybackslash}p{0.5000\atlwidth}>{\raggedright\arraybackslash}p{0.3600\atlwidth}@{}}
\caption{Capability-evidence ladder. Scores describe the available evidence for the declared target, not intelligence or learning gain.}\label{tab:P045}\\
\toprule
\textbf{Level} & \textbf{Evidence} & \textbf{Interpretation}\\
\midrule
\endfirsthead
\multicolumn{3}{l}{\small Table \thetable\ (continued)}\\
\toprule
\textbf{Level} & \textbf{Evidence} & \textbf{Interpretation}\\
\midrule
\endhead
\midrule
\multicolumn{3}{r}{\footnotesize Continued on next page}\\
\endfoot
\bottomrule
\endlastfoot
C=0 & No attributable demonstration beyond the artifact. & The evidence on record falls below either threshold. Ability itself remains open.\\[3pt]
C=1 & Self-report, generic reflection or a rehearsed account without probing. & Evidence of engagement, not adequate demonstration of the target capability.\\[3pt]
C=2 & Successful structured explanation of central reasoning or major revisions under declared aids. & Potentially adequate for a bounded, low-stakes formative objective.\\[3pt]
C=3 & Successful application, diagnosis or transfer on an appropriately controlled task, with the target reasoning attributable to the person. & Stronger evidence for the declared task; retention and growth need their own comparisons.\\[3pt]
\end{longtable}
\endgroup

For $C=3$, the person must complete the relevant task against the stated criteria, including when the task is supervised. If the reviewer establishes that the record contains no attributable demonstration, assign $C=0$. If the relevant evidence is unavailable and the reviewer cannot assign a score, record $U$. Set the context threshold $\tau(\Gamma)$ before review. The proposed threshold is 2 for a bounded formative explanation and 3 for independent application or readiness for a consequential task. Grading, selection and certification each require evidence of validity for that use and institutional authorization.

For an applicable capability claim with a known score,
\begin{equation}\label{eq:capability}
\operatorname{CapRes}=1\quad\Longleftrightarrow\quad C\geq\tau(\Gamma).
\end{equation}

Report the seven ratings, the gate decision, the status of each material claim, any applicable capability evidence and the revisions needed. Acceptance applies to the version, use and evidence reviewed. A confirmed deficiency requires revision. If the evidence needed for a decision is missing, leave that decision open. A tool-only review can later be extended to a learning claim by obtaining evidence of the designated learner's capability.

\Needspace{6\baselineskip}\section{Learning through inquiry and the resources it creates}

\Needspace{4\baselineskip}\subsection{Scientific inquiry: periodic order and local chaos in heart muscle}

In my research on cultured neonatal rat heart muscle cells, I studied sarcomeres, the contractile units arranged in series inside each cell. When the cells are warmed under specified conditions, the sarcomeres shorten and lengthen rapidly. These cycles are called hyperthermal sarcomeric oscillations (HSOs) \citep{ShintaniHSO2015}. A sarcomere can keep nearly the same period even as the distance it shortens varies substantially, a property called contraction rhythm homeostasis \citep{ShintaniHSO2020}. Measurements of neighbouring sarcomeres showed that their degree of synchrony changed and their local dynamics exhibited chaotic features \citep{ShintaniHSO2022}. I examined this coexistence of periodic order and local chaos using recurrence analysis, Lyapunov exponents and surrogate-data tests, and described it as \emph{chaordic homeodynamics} \citep{ShintaniHSO2025}.

To explore the recorded oscillations, I used symbolic regression, a method that searches for equations describing relations in data. The candidate equations prompted me to look beyond the motion of one sarcomere and ask how the timing of its contraction relates to that of its neighbour \citep{ShintaniKaken2023,ShintaniFitness2026}. AI helped me formulate that question. The evidence for the biological interpretation came from paired measurements and nonlinear analyses.

One length trace shows how far and how often a sarcomere shortens. Comparing two traces also shows whether the neighbours shorten together and whether their timing relationship persists from cycle to cycle. This change in what I look for in a recording is something I can carry into the next investigation. The measurement and analysis methods can also be used by other researchers with appropriate access.

\Needspace{4\baselineskip}\subsection{A research model: from fit to a bounded claim}\label{sec:binding}

Consider a molecule that reversibly binds to a site, such as a receptor. Suppose two sets of binding and dissociation rates fit every measurement made after the system has settled to equilibrium. A researcher who has used AI to fit the curve can make one more measurement. Should the researcher change the concentration and wait for equilibrium, or measure before equilibrium is reached? Consider one class of independent binding sites, a fixed free-ligand concentration $L$, and association and dissociation governed by mass action. The fraction of occupied sites, $\theta$, obeys
\begin{equation}\label{binding}
\frac{d\theta}{dt}=k_{\mathrm{on}}L(1-\theta)-k_{\mathrm{off}}\theta,
\end{equation}

\begin{equation}\label{equilibrium}
\theta_{\infty}(L)=\frac{L}{K_d+L},\qquad K_d=\frac{k_{\mathrm{off}}}{k_{\mathrm{on}}}.
\end{equation}

\begin{figure}[htbp]
\centering
\includegraphics[width=0.95\linewidth]{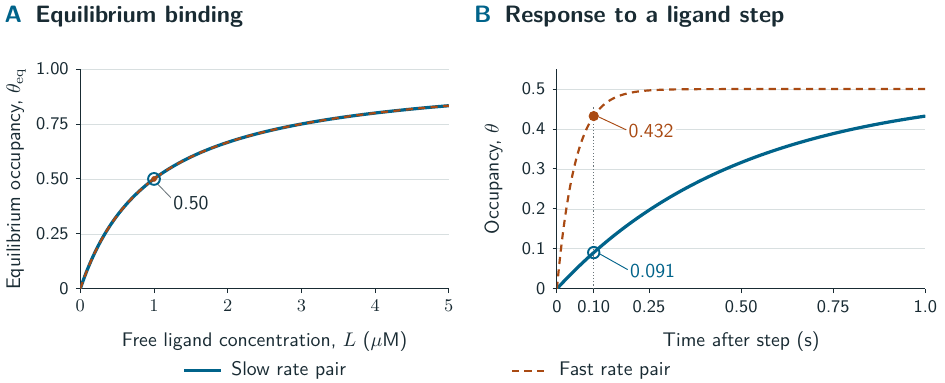}
\caption{Two binding models agree at equilibrium and separate in time. (A) Their rate pairs share $K_d=1\,\mu\mathrm{M}$, so their equilibrium curves coincide throughout the concentration range. (B) Starting from zero occupancy, a step to $L=1\,\mu\mathrm{M}$ free ligand produces relaxation times of 0.50 and 0.05~s. Both models approach 0.5. At 0.10~s their calculated occupancies are 0.091 and 0.432, respectively.}
\label{fig:binding}
\end{figure}

Multiplying both rate constants by the same positive factor leaves $K_d$ and every equilibrium prediction unchanged. Equilibrium data therefore cannot identify the common rate scale. For $L=1\,\mu\mathrm{M}$, the two pairs $(k_{\mathrm{on}},k_{\mathrm{off}})=(1,1)$ and $(10,10)$, in $\mu\mathrm{M}^{-1}\mathrm{s}^{-1}$ and $\mathrm{s}^{-1}$ respectively, each give an equilibrium occupancy of 0.5. Their relaxation times, $1/(k_{\mathrm{on}}L+k_{\mathrm{off}})$, are 0.50 and 0.05~s. Starting from zero occupancy, the predicted occupancies 0.10~s after the ligand step are 0.091 and 0.432 (Fig.~\ref{fig:binding}). A measurement during the approach to equilibrium can therefore distinguish these two candidates if its timing and precision resolve the difference.

I would keep both the calculation and the reason for choosing a transient measurement. In the five-part package, A contains the equations and their implementation, P the derivation and checks, and V the model's assumptions. F lists observations that would prompt me to reconsider the model, such as ligand depletion, transport effects or evidence for different classes of sites. R includes the script and its units. These materials let me revisit the choice, or a colleague check it, even if AI suggested the measurement or wrote the code.

The constructed cases B0 and B1 use the same equations, derivation, designated reviewer, nonsensitive synthetic inputs and files. They differ in how they document V and F. B0 offers a general caution. B1 names the limits of the model and observations that would warrant reconsideration.

\Needspace{4\baselineskip}\subsection{A learning task: improving evidence without changing the artifact}

The constructed cases L0 and L1 keep the checked explanation and its supporting records fixed. The sources, reasoning, scope, corrections, information handling and reuse instructions all meet the stipulated level-3 artifact anchors. In L0, the learner says they understood the explanation. In L1, the learner is assumed to explain the reasoning and apply it on a new task under declared aids. The change from $C=1$ to $C=3$ records the difference in the evidence supplied in these two hypothetical cases.

A transfer task can use the same equations while changing the starting condition. Give the learner this prompt: \emph{Both preparations are already at equilibrium at $L=1\,\mu\mathrm{M}$. A colleague proposes recording occupancy for the next 0.10~s with the bath unchanged. Could this distinguish the two rate pairs? If needed, change the procedure and explain when to measure and what difference to expect.} Provide the two rate pairs and their units from Section~\ref{sec:binding} with the prompt. The learner may consult the model equations and use a calculator.

For the assessor, an unchanged bath keeps both preparations at an occupancy of 0.5. One way to separate them is to wash out the free ligand and hold its concentration at zero. Then $\theta(t)=0.5\exp(-k_{\mathrm{off}}t)$, and the preparation with the larger off-rate has lower occupancy at any finite time after washout. Credit any answer that explains why the original plan fails and proposes a change that would separate the predictions with feasible timing and precision. A learner who has seen this solution needs a fresh task. No learner was assessed in this example.

\begingroup
\small
\setlength{\tabcolsep}{4pt}
\setlength{\atlwidth}{\dimexpr\textwidth-8\tabcolsep\relax}
\begin{longtable}{@{}>{\raggedright\arraybackslash}p{0.2300\atlwidth}>{\raggedright\arraybackslash}p{0.2200\atlwidth}>{\raggedright\arraybackslash}p{0.1400\atlwidth}>{\raggedright\arraybackslash}p{0.1600\atlwidth}>{\raggedright\arraybackslash}p{0.2500\atlwidth}@{}}
\caption{Worked scores for five constructed cases. The evidence records are supplied in the ancillary files.}\label{tab:P060}\\
\toprule
\textbf{Case} & \textbf{m1--m7} & \textbf{Gate} & \textbf{Capability} & \textbf{Action}\\
\midrule
\endfirsthead
\multicolumn{5}{l}{\small Table \thetable\ (continued)}\\
\toprule
\textbf{Case} & \textbf{m1--m7} & \textbf{Gate} & \textbf{Capability} & \textbf{Action}\\
\midrule
\endhead
\midrule
\multicolumn{5}{r}{\footnotesize Continued on next page}\\
\endfoot
\bottomrule
\endlastfoot
B0: Generic caution & 3,3,3,1,3,3,3 & Not met & Not claimed & Add V and F before the declared handoff.\\[3pt]
B1: Bounded model & 3,3,3,3,3,3,3 & Met & Not claimed & Retain the explicitly conditional model claim.\\[3pt]
L0: Self-report only & 3,3,3,3,3,3,3 & Met & C=1; target 3 & Demonstrate explanation and transfer.\\[3pt]
L1: Direct demonstration & 3,3,3,3,3,3,3 & Met & C=3; target 3 & Support the declared task-level capability claim.\\[3pt]
U1: Unchecked data handling & 3,3,3,3,U,3,3 & Not established & Not claimed & Obtain the missing information-handling evidence.\\[3pt]
\end{longtable}
\endgroup

Table~\ref{tab:P060} shows what each reviewer would do next. B0 requires a statement of scope and a response to failure. Both L0 and L1 pass the artifact review, but only L1 meets the predeclared capability threshold. In U1, the missing record of data handling leaves the assessment unresolved. Each case identifies the claim, evidence or anchor that another reviewer could question.

\Needspace{4\baselineskip}\subsection{An educational system built with AI assistance}

I developed a web application with generative AI to teach students preparing for the Japanese national examination for clinical laboratory technologists. Its question bank covers 21 annual examinations. I identified concepts that recur across years and prepared explanations for them. The questions, explanations, answer keys, feedback and practice rules are stored in editable form. The application uses those stored resources for routine search, practice, scoring and feedback without calling generative AI.

A student can begin with an examination year or topic, or search for a question by keyword. After answering, the student can open a linked topic explanation, try a check item and read the reasoning for each option. Questions from different years can therefore be studied together when they address the same concept. The system uses the student's response history to present missed questions again.

Instructors can examine which topics recur in their subject area and which incorrect options students tend to select. They can use those patterns to choose a concept to revisit in class or an explanation to revise. As new examinations become available or a problem is identified, I can update the materials and practice rules, with further AI assistance when useful.

The system gives a student a way to follow one concept across several questions and gives an instructor a place to revise the explanation once a difficulty comes to light. The distinction between practice and learning matters here. A student who can answer a familiar item on a second attempt could next be asked to explain the same concept in an unfamiliar problem. Such a task could provide evidence of transfer. The learning effects of this system remain to be measured.

\Needspace{6\baselineskip}\section{Learning, autonomy and the accumulation of knowledge}

\subsection{Making each inquiry a starting point for the next}

A researcher can ask AI for candidate models, then examine their assumptions and choose a measurement that would distinguish them. A student can work through a suggested solution, explain the principle and try it on an unfamiliar problem. Keeping the derivation, the program or a corrected explanation gives each person a point of departure when a harder question arises. The reasoning they can now do for themselves shapes what they ask of AI next.

The time spent on checks should reflect the goal and the consequences of error. For a study note, a source, a worked example and a short explanation may be enough. For a laboratory script used in later analyses, it may be worth documenting the environment, specifying a test and assigning responsibility for maintenance. As tools improve, the skill to practise may shift from implementing a calculation by hand to choosing its assumptions or diagnosing a discrepancy.

\subsection{Choosing what to delegate and what to retain}

Delegating an operation need not mean giving up control of the knowledge that guides it. A laboratory might ask an external tool to write a plotting routine using example values, then run it on unpublished measurements within its own environment. The group can keep the analysis, its assumptions and the reasons for its choices together. It can decide which materials to retain privately, which an authorized colleague may inspect, and which to publish. Permissions to process, store, share or train on data are distinct \citep{Autio2024}. A published claim or regulated use must also be supported by evidence that the appropriate reviewers can examine.

A group can accumulate expertise when its members have both a working method and opportunities to exercise the judgment the method requires. A new member of a laboratory might rerun an analysis and then decide how to adapt it to a different instrument. A new instructor might revise a teaching explanation after examining students' errors. Over time, teaching such judgments alongside the materials that support them offers one way for institutions and societies to pass knowledge across generations.

\subsection{Increasingly capable AI and the value of understanding}

A mathematical proof illustrates the distinction even when AI finds a solution beyond a learner's present skill. Given a formal statement, its assumptions and an appropriate checking environment, a proof assistant can check that a derivation establishes the statement. The learner can take the proof further by identifying where a hypothesis enters the argument or adapting a step when that hypothesis changes. Both the checked proof and the ability to reason about it have value.

In empirical work, a calculation and a test can agree because both encode the same mistaken assumption. An independent derivation, a different measurement or a comparison with a rival model can bring the mistake to light. Increasing the reliability of AI changes which checks are worth performing and which skills are worth practising. The choice of what a person or a group wants to understand still matters. The review instrument in Section~\ref{sec:artifact} covers retained artifacts whose declared routine use is independent of the original AI. A system that calls that AI during routine use would require additional criteria.

\Needspace{6\baselineskip}\section{Verification, limitations and a validation plan}

\Needspace{4\baselineskip}\subsection{What the executable checks establish}

The binding program reproduces the equilibrium and transient predictions. The equations explain their difference: rescaling both rate constants preserves their ratio but changes the relaxation time. The rubric software takes a different kind of input, the ratings and evidence judgments supplied by a reviewer.

I tested the rule for combining those ratings by increasing one score at a time across all fully rated profiles. Every increase preserved an existing gate pass. Some profiles with a low total passed while others with a higher total failed because one critical rating fell below threshold. In the constructed records, new adverse evidence could reverse a decision, while changing the capability score alone left the artifact profile unchanged. The English and Japanese offline sheets agreed with the Python implementation on every tested profile and assessment record. Appendix Section~\ref{sec:executable} gives the inputs, commands and counts.

\Needspace{4\baselineskip}\subsection{Empirical validation}

The scoring anchors and thresholds are provisional. The program can apply the same rule consistently to a given set of ratings. Whether independent reviewers assign similar ratings to the same evidence, and whether their decisions help people correct or reuse an artifact, require observations in practice.

Review also takes time and may expose confidential information. A level-4 rating records a check performed within the stated scope. Its meaning depends on the check: a derivation can establish a property of a model, while observations can test how well the model describes a particular system. Where access is restricted, record what an authorized reviewer can inspect and what evidence is publicly available.

\Needspace{4\baselineskip}\subsection{An evaluation that could improve or reject the instrument}\label{sec:validation}

My first priority would be to test whether diagnosing and correcting an error helps learners apply a concept to a new problem. Both groups could use the same physiology materials for the same amount of time. One group would diagnose and correct an erroneous explanation before seeing a worked correction. The comparison group would study the worked correction. I would specify baseline capability, group allocation, permitted aids, scoring criteria and a common unseen task about a week later. The study would measure performance with assistance, immediate performance attributable to the learner, and delayed transfer separately. To study retention, it would also reassess the original capability later. Random assignment could estimate the effect of the correction activity relative to studying the worked correction. An index of the divergence between performance and learning, considered in v1, would require comparable scales and estimates of measurement error.

I would ask intended users to apply the anchors to incomplete, borderline and well-supported packages. Independent reviewers would rate the same evidence without knowing the intended ratings. The analysis would identify which anchors invite different readings and report agreement and uncertainty for each ordinal dimension and for the gate decision. Methods for measuring multi-observer ordinal agreement must match the rating design \citep{Vanbelle2024}. An overall measure of internal consistency cannot establish the validity of seven deliberately distinct requirements.

A handoff study could give independent maintainers two versions of a research package. Both would contain code, a README, the environment specification and tests. One would present the evidence in a conventional README, the other under the five proposed elements. Matching content and length would help isolate the effect of organization. Maintainers would locate and correct a seeded error, run an independent check and adapt the result to a new condition. The expected correction, success criteria, time limit and permitted assistance would be set in advance. Outcome assessors could be blinded to the package condition and the task order counterbalanced. Endpoints would include correction accuracy, new errors, time, requests for clarification and review effort. Little improvement in handoff or correction, or a disproportionate burden, would weigh against adoption. A separate trial could compare the full package with ordinary practice.

In a \emph{dependency-switch test}, maintainers would perform matched tasks under three conditions: the original AI is available, a different AI is available, or the original AI is unavailable and no substitute is provided. With the same files, time and ordinary tools, the comparison would show which steps depend on continued access to a particular AI. The protocol would record tool versions, operations material to the result, costs and outcomes. Synthetic errors could be used first to check that the tasks can reveal the intended dependence.

\Needspace{6\baselineskip}\section{Conclusion}

Learning with AI can begin with a result that lies beyond one's present skill. I propose asking what that result now makes it possible to understand, and what evidence would show that the person can use the understanding again. A saved program or explanation gives the next inquiry a starting point. A demonstration under stated conditions shows what the person can do with what they learned. The proposed review method records each kind of evidence separately.

In my research on heart muscle, a candidate equation led me to ask how neighbouring contractile units oscillate together. In the binding example, understanding why two models agree at equilibrium guides the choice of a measurement that separates them. In examination practice, students can return to the concepts behind missed questions, and I can use patterns of incorrect responses to improve teaching materials. The decision rules have been implemented and checked. Whether using them improves learning or helps others revise an artifact remains to be tested in practice.

What I want to carry from one inquiry to the next is both a resource I can use and a judgment I have learned to make. In a research group or a classroom, keeping the reasoning with the resource gives others an opportunity to learn that judgment, test it under new conditions and pass on what they discover. This is how learning with AI could contribute to the accumulation of knowledge across generations.

\section*{Declarations}

\subsection*{Funding}

This work was supported by JSPS KAKENHI Grant Number JP25K00269; the Chubu University FY2025 Special Research Fund (CP), ``Development of an AI safety evaluation and AI to Learn-based utilization system for clinical laboratory technologist education''; and the Chubu University FY2025 Research Institute for Industry and Economics Research Project, ``Development and evaluation of an explainable management-analysis support tool using local small-scale LLMs.''

\subsection*{Competing interests and author contribution}

I declare no financial competing interests. I conceived AI to Learn, developed the framework and educational system, prepared the cases and accompanying materials, and wrote and revised the manuscript.

\subsection*{Ethics, data and code availability}

No participants were recruited or assessed and no individual learner records were analyzed for this paper. The sarcomere case draws on published experiments and analyses. The manuscript includes the full rubric, constructed review records, analytical example and verification programs. The educational application's source code and institutional data are not distributed with this paper. The proposed studies with participants require appropriate institutional review and consent.

\subsection*{AI-assisted preparation}

I used ChatGPT and Codex to assist with drafting, translation, reference checks, and preparation and testing of the accompanying code and materials. I take responsibility for the arguments, sources and final content.

\begingroup
\small
\setlength{\bibsep}{0pt}
\bibliographystyle{unsrtnat}
\bibliography{references}
\endgroup
\clearpage
\section*{Technical appendix}
\addcontentsline{toc}{section}{Technical appendix}
\setcounter{section}{0}
\setcounter{table}{0}
\renewcommand{\thesection}{S\arabic{section}}
\renewcommand{\thetable}{S\arabic{table}}
\renewcommand{\theHsection}{S\arabic{section}}
\renewcommand{\theHtable}{S\arabic{table}}
\setlength{\LTpre}{4pt}\setlength{\LTpost}{4pt}
\Needspace{6\baselineskip}\section{Full scoring anchors}\label{sec:anchors}

Rate the declared deliverable for its intended purpose. The scale runs from an established absence of the practice (0), through a weak or asserted practice (1) and partial practice (2), to documented sufficiency (3) and sufficiency demonstrated by a test or use (4). Apply the specific criterion for each dimension. If the evidence needed to choose a rating is unavailable, record U. Give a reason and an evidence location for each rating. Assess any claim about human capability separately.

\Needspace{16\baselineskip}\subsection{m1: Residual opacity}

\begingroup
\small
\setlength{\tabcolsep}{4pt}
\setlength{\atlwidth}{\dimexpr\textwidth-2\tabcolsep\relax}
\begin{longtable}{@{}>{\raggedright\arraybackslash}p{0.0900\atlwidth}>{\raggedright\arraybackslash}p{0.9100\atlwidth}@{}}
\caption{Independence from the originating AI in use and justification.}\label{tab:P098}\\
\toprule
\textbf{Score} & \textbf{Anchor}\\
\midrule
\endfirsthead
\multicolumn{2}{l}{\small Table \thetable\ (continued)}\\
\toprule
\textbf{Score} & \textbf{Anchor}\\
\midrule
\endhead
\midrule
\multicolumn{2}{r}{\footnotesize Continued on next page}\\
\endfoot
\bottomrule
\endlastfoot
0 & The stated result must be generated again by the original opaque AI, or its important claims are accepted solely on that AI's assurance.\\[3pt]
1 & A static output is retained, but core operations or reasons remain recoverable only by returning to the AI.\\[3pt]
2 & Some operations or grounds are explicit; at least one material dependency on opaque generation remains unresolved.\\[3pt]
3 & Core use and justification do not require the originating opaque AI; the relevant evidence is accessible and noncritical gaps are stated.\\[3pt]
4 & An appropriate reviewer or test has exercised use, inspection and revision without that AI, and the demonstrated scope is recorded.\\[3pt]
\end{longtable}
\endgroup

\Needspace{16\baselineskip}\subsection{m2: Human authority}

\begingroup
\small
\setlength{\tabcolsep}{4pt}
\setlength{\atlwidth}{\dimexpr\textwidth-2\tabcolsep\relax}
\begin{longtable}{@{}>{\raggedright\arraybackslash}p{0.0900\atlwidth}>{\raggedright\arraybackslash}p{0.9100\atlwidth}@{}}
\caption{Authority to accept, correct or withdraw the work.}\label{tab:P100}\\
\toprule
\textbf{Score} & \textbf{Anchor}\\
\midrule
\endfirsthead
\multicolumn{2}{l}{\small Table \thetable\ (continued)}\\
\toprule
\textbf{Score} & \textbf{Anchor}\\
\midrule
\endhead
\midrule
\multicolumn{2}{r}{\footnotesize Continued on next page}\\
\endfoot
\bottomrule
\endlastfoot
0 & No accountable person or team can accept, correct or withdraw the result.\\[3pt]
1 & A nominal approver is named, but practical authority or criteria are unclear.\\[3pt]
2 & Meaningful review occurs, but release, correction or withdrawal responsibilities are incompletely specified.\\[3pt]
3 & Responsible roles, acceptance criteria and feasible release, correction and withdrawal routes are documented.\\[3pt]
4 & The documented arrangement has been exercised through an actual revision, rejection or a clearly labeled test of the intervention route.\\[3pt]
\end{longtable}
\endgroup

\Needspace{16\baselineskip}\subsection{m3: Traceability}

\begingroup
\small
\setlength{\tabcolsep}{4pt}
\setlength{\atlwidth}{\dimexpr\textwidth-2\tabcolsep\relax}
\begin{longtable}{@{}>{\raggedright\arraybackslash}p{0.0900\atlwidth}>{\raggedright\arraybackslash}p{0.9100\atlwidth}@{}}
\caption{Tracing a claim to its sources, decisions and checks.}\label{tab:P102}\\
\toprule
\textbf{Score} & \textbf{Anchor}\\
\midrule
\endfirsthead
\multicolumn{2}{l}{\small Table \thetable\ (continued)}\\
\toprule
\textbf{Score} & \textbf{Anchor}\\
\midrule
\endhead
\midrule
\multicolumn{2}{r}{\footnotesize Continued on next page}\\
\endfoot
\bottomrule
\endlastfoot
0 & No usable record of the sources, operations or checks material to the accepted claims is retained.\\[3pt]
1 & There are only unsupported assertions about material sources or verification; the evidential route cannot be reconstructed.\\[3pt]
2 & Some sources and checks are available, but a material data transformation, selection decision or verification step remains unclear.\\[3pt]
3 & Sources, relevant tool roles, data transformations, selection decisions, substantive revisions and verification can be reconstructed for the accepted result.\\[3pt]
4 & An appropriate independent reviewer has used versioned records and checks to reconstruct or restabilize the accepted result within the declared scope.\\[3pt]
\end{longtable}
\endgroup

\Needspace{16\baselineskip}\subsection{m4: Validity and failure response}

\begingroup
\small
\setlength{\tabcolsep}{4pt}
\setlength{\atlwidth}{\dimexpr\textwidth-2\tabcolsep\relax}
\begin{longtable}{@{}>{\raggedright\arraybackslash}p{0.0900\atlwidth}>{\raggedright\arraybackslash}p{0.9100\atlwidth}@{}}
\caption{Defining the scope and responding to failure.}\label{tab:P104}\\
\toprule
\textbf{Score} & \textbf{Anchor}\\
\midrule
\endfirsthead
\multicolumn{2}{l}{\small Table \thetable\ (continued)}\\
\toprule
\textbf{Score} & \textbf{Anchor}\\
\midrule
\endhead
\midrule
\multicolumn{2}{r}{\footnotesize Continued on next page}\\
\endfoot
\bottomrule
\endlastfoot
0 & V: no intended scope or assumptions. F: no response to error or contradiction.\\[3pt]
1 & V: only a generic description. F: only ``use with caution'' or an equivalent disclaimer.\\[3pt]
2 & V: some assumptions or exclusions, with material boundaries missing. F: some failure cases, but triggers or response routes are incomplete.\\[3pt]
3 & V: purpose, assumptions, applicable regime and exclusions are explicit. F: observable reconsideration triggers and correction or withdrawal routes are specified.\\[3pt]
4 & V: relevant boundary or alternative cases have been examined. F: negative tests or a documented correction exercise have tested the response within the declared scope.\\[3pt]
\end{longtable}
\endgroup

\Needspace{16\baselineskip}\subsection{m5: Information protection}

\begingroup
\small
\setlength{\tabcolsep}{4pt}
\setlength{\atlwidth}{\dimexpr\textwidth-2\tabcolsep\relax}
\begin{longtable}{@{}>{\raggedright\arraybackslash}p{0.0900\atlwidth}>{\raggedright\arraybackslash}p{0.9100\atlwidth}@{}}
\caption{Protecting information during the work.}\label{tab:P106}\\
\toprule
\textbf{Score} & \textbf{Anchor}\\
\midrule
\endfirsthead
\multicolumn{2}{l}{\small Table \thetable\ (continued)}\\
\toprule
\textbf{Score} & \textbf{Anchor}\\
\midrule
\endhead
\midrule
\multicolumn{2}{r}{\footnotesize Continued on next page}\\
\endfoot
\bottomrule
\endlastfoot
0 & Information requiring protection is exposed without appropriate controls, or permissions materially exceed the justified task.\\[3pt]
1 & Informal caution or ad hoc redaction is described, but relevant data flows and access are not controlled.\\[3pt]
2 & Some minimization or access controls exist, but material recipients, uses or retention conditions remain unresolved.\\[3pt]
3 & Relevant data classes, recipients, permitted uses, access, minimization and retention are documented with task-appropriate controls; exclusively synthetic/public inputs are explicitly recorded where applicable.\\[3pt]
4 & The controls have been examined through a proportionate data-flow, access, deletion or equivalent protection check; the check and its limits are recorded.\\[3pt]
\end{longtable}
\endgroup

\Needspace{16\baselineskip}\subsection{m6: Resource proportionality}

\begingroup
\small
\setlength{\tabcolsep}{4pt}
\setlength{\atlwidth}{\dimexpr\textwidth-2\tabcolsep\relax}
\begin{longtable}{@{}>{\raggedright\arraybackslash}p{0.0900\atlwidth}>{\raggedright\arraybackslash}p{0.9100\atlwidth}@{}}
\caption{Resources needed to sustain the intended use.}\label{tab:P108}\\
\toprule
\textbf{Score} & \textbf{Anchor}\\
\midrule
\endfirsthead
\multicolumn{2}{l}{\small Table \thetable\ (continued)}\\
\toprule
\textbf{Score} & \textbf{Anchor}\\
\midrule
\endhead
\midrule
\multicolumn{2}{r}{\footnotesize Continued on next page}\\
\endfoot
\bottomrule
\endlastfoot
0 & Critical resources or access are unavailable, or continuing requirements have no credible relation to the intended use.\\[3pt]
1 & Dependencies are acknowledged but costs, access or interruption consequences are unmanaged.\\[3pt]
2 & Some requirements or reductions are described, but ordinary operation or recovery remains under-specified.\\[3pt]
3 & Routine software, access, relevant cost estimates and interruption responses are feasible and documented; estimates are distinguished from measurements.\\[3pt]
4 & Operation or recovery has been exercised against documented requirements, and alternatives or resource trade-offs have been examined with an explicit basis.\\[3pt]
\end{longtable}
\endgroup

\Needspace{16\baselineskip}\subsection{m7: Reuse and handoff}

\begingroup
\small
\setlength{\tabcolsep}{4pt}
\setlength{\atlwidth}{\dimexpr\textwidth-2\tabcolsep\relax}
\begin{longtable}{@{}>{\raggedright\arraybackslash}p{0.0900\atlwidth}>{\raggedright\arraybackslash}p{0.9100\atlwidth}@{}}
\caption{Reusing and revising the work from its retained materials.}\label{tab:P110}\\
\toprule
\textbf{Score} & \textbf{Anchor}\\
\midrule
\endfirsthead
\multicolumn{2}{l}{\small Table \thetable\ (continued)}\\
\toprule
\textbf{Score} & \textbf{Anchor}\\
\midrule
\endhead
\midrule
\multicolumn{2}{r}{\footnotesize Continued on next page}\\
\endfoot
\bottomrule
\endlastfoot
0 & Only the original creator can meaningfully operate or interpret the retained work.\\[3pt]
1 & Another person could proceed only with extensive direct help; essential instructions or files are absent.\\[3pt]
2 & Partial instructions permit some reuse, but a material task still depends on unrecorded creator knowledge.\\[3pt]
3 & An appropriate person has the instructions, editable materials and prerequisites needed for the declared use and modification without creator recollection.\\[3pt]
4 & Another appropriate person has completed a relevant use, modification or restoration task from the package, with the outcome and remaining limits recorded.\\[3pt]
\end{longtable}
\endgroup

Rate the V and F components of m4 separately. When both are known, $m_4=\min(m_V,m_F)$. If one rating is unknown and the other is below 3, the condition is \emph{not met}. If a rating is unknown and neither component is known to be below 3, the condition is \emph{not established}. A workflow using only public or synthetic inputs can meet m5 when the input restriction and relevant controls are documented. For m6, comparisons of energy use require a stated system boundary and measurements or justified estimates. Running a model locally or avoiding model calls does not by itself supply that evidence.

\Needspace{6\baselineskip}\section{Reusable review record}

Use the accompanying Markdown template or JSON schema to record the review. The schema specifies required fields and permissible values. The reviewer judges whether the evidence is accurate, relevant and sufficient. For records intended for publication, describe how authorized reviewers can inspect restricted evidence and keep private learner responses and confidential data out of the public record.

\Needspace{15\baselineskip}
\begingroup
\small
\setlength{\tabcolsep}{4pt}
\setlength{\atlwidth}{\dimexpr\textwidth-2\tabcolsep\relax}
\begin{longtable}{@{}>{\raggedright\arraybackslash}p{0.2200\atlwidth}>{\raggedright\arraybackslash}p{0.7800\atlwidth}@{}}
\caption{Record fields and their function.}\label{tab:P114}\\
\toprule
\textbf{Field} & \textbf{Required content}\\
\midrule
\endfirsthead
\multicolumn{2}{l}{\small Table \thetable\ (continued)}\\
\toprule
\textbf{Field} & \textbf{Required content}\\
\midrule
\endhead
\midrule
\multicolumn{2}{r}{\footnotesize Continued on next page}\\
\endfoot
\bottomrule
\endlastfoot
Context & Task, artifact version, purpose, users, review point, material AI roles, and capability claim if applicable.\\[3pt]
Claim and evidence & Exact claim; acceptance criteria; evidence location; supported, contradicted or not-established status; key limitations.\\[3pt]
D package & Locations of A, P, V, F and R; a human judgment of material sufficiency for the declared purpose.\\[3pt]
M profile & Each score or U; evidence locator; justification; reviewer and date.\\[3pt]
Capability & Target task and success criteria, allowed aids, timing, result, C and predeclared threshold; no personal records in public examples.\\[3pt]
Decision & Gate status; artifact and capability conclusions kept separate; positive scope or reason to revise or defer.\\[3pt]
Improvement & Specific action, responsible role, required new evidence and next review point.\\[3pt]
\end{longtable}
\endgroup

State the question and any capability claim, assemble the artifact and its evidence, then assign ratings. For a confirmed deficiency, specify the correction needed. For missing evidence, identify what would permit a decision. Reassess the work when the purpose or version changes or when adverse evidence emerges.

\Needspace{6\baselineskip}\section{Constructed records and reproducible checks}\label{sec:executable}

The five records in \path{anc/examples.json} are synthetic. Their common level-3 evidence consists of an editable artifact, a named review role, records of sources, tool use and checks, public or synthetic inputs with explicit limits, documented operational resources and reuse instructions. B0 gives a generic warning for V and F, while B1 states the relevant scope and reasons to reconsider the model. L0 and L1 use the same artifact. L0 assumes a self-report and L1 assumes successful explanation and transfer. In U1 the reviewer lacks evidence about information handling and records both m5 and the sufficiency of the package as unknown.

Run \texttt{python3} \path{anc/review_rules.py} \texttt{-{}-self-test} to enumerate fully rated profiles and test the decisions. Run \texttt{python3} \path{anc/review_rules.py} \path{anc/examples.json} for the five case reports. JSON records use integers from 0 to 4 for known artifact ratings and \texttt{null} for $U$. A known failure takes precedence over an unresolved critical rating. When package sufficiency is unknown, the program returns \emph{not established} for that component. A failed capability requirement affects the combined claim and leaves the artifact profile unchanged.

The English and Japanese sheets in \path{anc/review_sheet} can be opened offline. To compare their decisions with the Python implementation, run \texttt{python3} \path{anc/review_sheet/verify.py} with Node.js on the system path. Additional checks exercise record import and export, including omission of capability evidence when the reviewer turns off the capability claim.

Of the 78,125 fully rated seven-dimensional profiles, 2,000 passed the four-dimension gate. Across 437,500 possible one-step increases in a single rating, every profile that already passed continued to pass. One passing profile had a sum of 12, while one failing profile had a sum of 26. In each language, the browser implementation agreed with Python on all 279,936 profiles that include the unknown state among the possible ratings. The implementations also agreed on 131 assessment records covering the constructed examples and variations in V, F, capability, claim status and package sufficiency.

The program \path{anc/binding_example.py} uses the Python standard library to compute both models' equilibrium and transient solutions and recover their assumed rate constants. Starting at zero occupancy, $\theta(t)=\theta_\infty[1-\exp\{-(k_{\mathrm{on}}L+k_{\mathrm{off}})t\}]$. At $L=1\,\mu\mathrm{M}$ the calculated occupancies at 0.1 s are approximately 0.090635 and 0.432332. Under the same assumptions, the relaxation rate $\lambda$ gives $k_{\mathrm{on}}=\lambda/(L+K_d)$ and $k_{\mathrm{off}}=K_d\lambda/(L+K_d)$. A departure from the assumptions calls for reconsidering the inference.

\Needspace{6\baselineskip}\section{Validation tasks and continuity with v1}

In the learning study proposed in Section~\ref{sec:validation}, both groups would study the same fixed, AI-assisted physiology materials for the same amount of time. One group would diagnose and correct the erroneous explanation before reading the worked correction. The other group would study the worked correction. Both groups would receive the same unseen problem about a week later. Baseline capability, permitted aids, scoring criteria and group allocation would be specified in advance. Random assignment would allow an estimate of the effect of the correction activity under those conditions.

In the proposed reviewer study, use neutral case labels and define target defects independently of rubric scoring. Report agreement on the ordinal ratings by dimension, disagreements on the gate, the frequency of unknown ratings and review time. Compare decisions and correction efficiency with those obtained through ordinary disciplinary review of a conventional README.

\begingroup
\small
\setlength{\tabcolsep}{4pt}
\setlength{\atlwidth}{\dimexpr\textwidth-4\tabcolsep\relax}
\begin{longtable}{@{}>{\raggedright\arraybackslash}p{0.2000\atlwidth}>{\raggedright\arraybackslash}p{0.3200\atlwidth}>{\raggedright\arraybackslash}p{0.4800\atlwidth}@{}}
\caption{Operational continuity with the first 2.0 manuscript.}\label{tab:P123}\\
\toprule
\textbf{Component} & \textbf{Retained} & \textbf{Clarified or revised}\\
\midrule
\endfirsthead
\multicolumn{3}{l}{\small Table \thetable\ (continued)}\\
\toprule
\textbf{Component} & \textbf{Retained} & \textbf{Clarified or revised}\\
\midrule
\endhead
\midrule
\multicolumn{3}{r}{\footnotesize Continued on next page}\\
\endfoot
\bottomrule
\endlastfoot
Dual residuals & Separate artifact and human evidence. & A human checkpoint cannot repair an unchanged artifact; claims of retention or growth need additional evidence.\\[3pt]
Package & D=(A,P,V,F,R). & The retained artifact may differ from the AI proposal. The review follows the evidence supporting the accepted work.\\[3pt]
Seven dimensions & 0--4 anchored profile and four critical dimensions. & U distinguishes missing evidence; m4 uses the weaker of V/F; m6 is not a proxy for local hosting. m3 concerns the sources, operations and checks needed to examine the result.\\[3pt]
Gate and capability & Threshold-3 critical gate and context-dependent C ladder. & Provisional design defaults; no total-score certification bands or automatic credential decisions.\\[3pt]
Worked scoring & Contrasts that make decisions inspectable. & Synthetic evidence records for comparing decision rules.\\[3pt]
Verification & Explicit, reproducible decision logic. & Reference implementation, exhaustive gate checks and a separate analytical calculation.\\[3pt]
\end{longtable}
\endgroup

The present revision retains AI to Learn 2.0 and its focus on reusable work and attributable understanding. The criteria and executable rules provide a common basis for further empirical evaluation.

\paragraph{Correction to the published v1 figure.} Figure~2 of arXiv:2604.19751v1 labels C2 and C6 as \(C=1/3\). Table~4 of that version gives \(C=3\) and a required threshold \(\tau=3\) for both. The intended score/threshold display is therefore \(3/3\); the binary indicator of threshold attainment is 1. This annotation does not alter the totals, critical-gate decisions or overall decisions reported there.

\end{document}